\documentclass[a4paper,conference]{IEEEtran}
\ifCLASSOPTIONcompsoc
\usepackage[nocompress]{cite}
\else
\usepackage{cite}
\fi
\usepackage{amsmath}

\usepackage{graphicx}
\usepackage{multirow}
\usepackage{multicol}
\usepackage{booktabs}
\usepackage{subfigure}

\ifCLASSINFOpdf
\else
\fi
\usepackage{tikz}
\usetikzlibrary{positioning}
\usepackage[hyphens]{url}

\begin{document}
%
\title{Perceptual Crack Detection for Rendered 3D Textured Meshes}
\author{\IEEEauthorblockN{Armin Shafiee Sarvestani$^{1}$, Wei Zhou$^{2}$, Zhou Wang$^{1}$ \\
$^{1}$Dept. of Electrical \& Computer Engineering, University of Waterloo, Canada \\
$^{2}$School of Computer Science and Informatics, Cardiff University, United Kingdom \\
Email: \{a5shafie, zhou.wang\}@uwaterloo.ca \qquad zhouw26@cardiff.ac.uk}}


%


\IEEEoverridecommandlockouts

\maketitle

\begin{abstract}
Recent years have witnessed many advancements in the applications of 3D textured meshes. As the demand continues to rise, evaluating the perceptual quality of this new type of media content becomes crucial for quality assurance and optimization purposes. Different from traditional image quality assessment, crack is an annoying artifact specific to rendered 3D meshes that severely affects their perceptual quality. In this work, we make one of the first attempts to propose a novel Perceptual Crack Detection (PCD) method for detecting and localizing crack artifacts in rendered meshes. Specifically, motivated by the characteristics of the human visual system (HVS), we adopt contrast and Laplacian measurement modules to characterize crack artifacts and differentiate them from other undesired artifacts. Extensive experiments on large-scale public datasets of 3D textured meshes demonstrate effectiveness and efficiency of the proposed PCD method in correct localization and detection of crack artifacts. 
Moreover, to quantify the performance of the proposed detection method and validate its effectiveness, we propose a simple yet effective weighting mechanism to incorporate the resulting crack map into classical quality assessment (QA) models, which creates significant performance improvement in predicting the perceptual image quality when tested on public datasets of static 3D textured meshes. A software release of the proposed method is publicly available at: https://github.com/arshafiee/crack-detection-VVM 
\end{abstract}

\begin{tikzpicture}[overlay, remember picture]
\path (current page.north) node (anchor) {};
\end{tikzpicture}
%


%
\IEEEpeerreviewmaketitle

\vspace{-0.8cm}
\section{Introduction}
With the advancement of 3D acquisition technologies and 3D processing tools and displays, the interest in immersive media has grown substantially in recent years. 3D mesh is one of the most promising media forms for 3D content representation and demonstrates great potential in many real-world applications such as medical imaging/modeling, creative storytelling, social virtual reality, and video gaming \cite{palomar2023mr, young2023volumetric, li2023social}. A 3D polygonal mesh is defined by a set of vertices in the 3D space. In addition to the xyz coordinates of the vertices, connectivity information is needed to form polygons (typically triangles). To colorize such 3D structures, either (1) color values are defined for each vertex, in which case the mesh is called a \textit{vertex-color mesh}; or (2) a 2D texture map is provided separately along with the 3D data to form colored \textit{textured meshes}. Mapping information between the 3D model space and the 2D texture space (UV mapping information) is then included in the 3D data to help the rendering method unfold the 2D texture map onto the 3D colorless object \cite{alexiou2023subjective}. 

\begin{figure}[t]
	\centerline{\includegraphics[width=8.8cm]{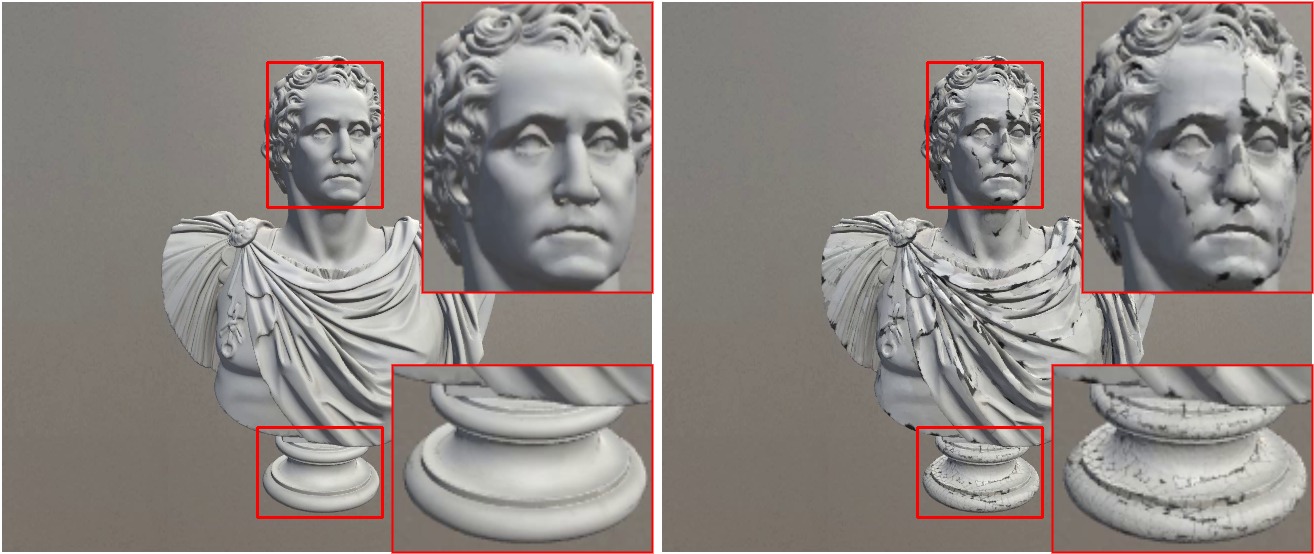}}
	\caption{Sample snapshots of a pair of reference (left) and distorted (right) 3D textured meshes. Crack artifacts are present all over the distorted object. Two local windows are cropped and enlarged for better visualization.}
	\centering
	\label{fig:crack_smp}
\end{figure}

3D textured meshes undergo a variety of processing operations including but not limited to simplification, quantization, compression, transmission, post-processing and rendering, throughout the multimedia supply chain. These operations will inevitably cause diverse distortions that degrade the perceived quality of the contents. Among those distortions, crack creates highly noticeable perceptual artifacts, as exemplified in Fig.~\ref{fig:crack_smp}. Crack artifacts are fractures or holes that appear on the surface of objects. They are specific to 3D meshes and are rarely observed in either distorted natural images/videos or other forms of 3D content. Different sources of distortions may cause crack artifacts depending on their strength level and the characteristics of the 3D content. However, crack artifacts appear mostly because of vertex position and UV quantization and drastically degrade the perceived quality of 3D objects. Consequently, a tool that can detect and localize crack artifacts is highly desirable and may be employed (1) to boost the performance of existing 3D quality assessment (QA) methods; and (2) to help optimize various mesh processing algorithms to detect, localize, reduce and rectify cracks.


Existing 3D mesh QA methods can be generally categorized as model-based and image-based metrics. While model-based metrics operate on the 3D object itself, image-based methods take 2D snapshots of the 3D objects as input \cite{alexiou2023subjective}. Many model-based \cite{aspert2002mesh, lavoue2006perceptually, lavoue2011multiscale, vavsa2012dihedral, wang2012fast, nehme2020visual} and image-based \cite{Abouelaziz2018, abouelaziz2020no, nehme2023textured, eep-3dqa} methods have been proposed for 3D mesh QA over the years, but to the best of our knowledge, no image-based or model-based method has been yet developed to detect and localize crack artifacts of textured meshes despite their severe impact on the perceived quality of 3D objects.

To address this issue, we propose an image-based Perceptual Crack Detection method (PCD) based on the characteristics of the human visual system (HVS) such as visual masking \cite{masking_eff, masking_eff2} and psychometric saturation \cite{wang2012fast} effects. Given a pair of snapshots of distorted and reference 3D objects, our proposed algorithm creates a crack likelihood map that indicates locations of cracks at pixel-level. Furthermore, to quantify the performance of the proposed PCD method and confirm its effectiveness, an efficient method is proposed to convert the crack map into a weight map, which is subsequently combined with quality maps generated by existing QA models to greatly enhance their performance in predicting the perceptual quality of the distorted meshes when tested using two large-scale subject-rated datasets.




\section{Proposed Method}


\subsection{Crack Detection} \label{ssec: crack_local}

\begin{figure}[t]
	\centerline{\includegraphics[width=8.8cm]{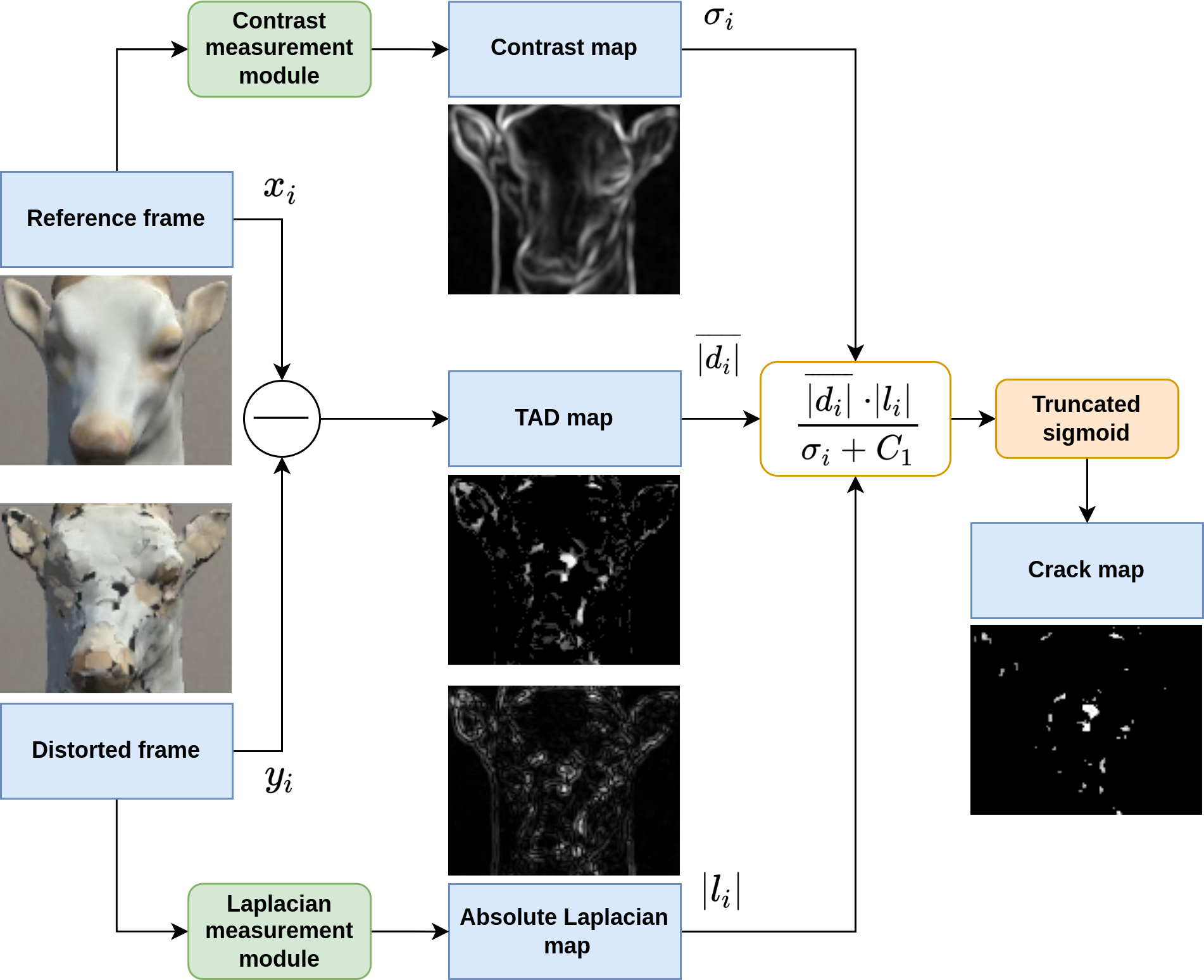}}
	\caption{PCD framework, where TAD represents the truncated absolute difference. The sample reference and distorted frames are obtained from Nehm{\'e} \emph{et al.} dataset \cite{nehme2023textured}.}
	\centering
	\label{fig:crack_detect}
\end{figure}



    

\begin{figure*}
    \centering

    \centerline{\includegraphics[width=1.0\textwidth]{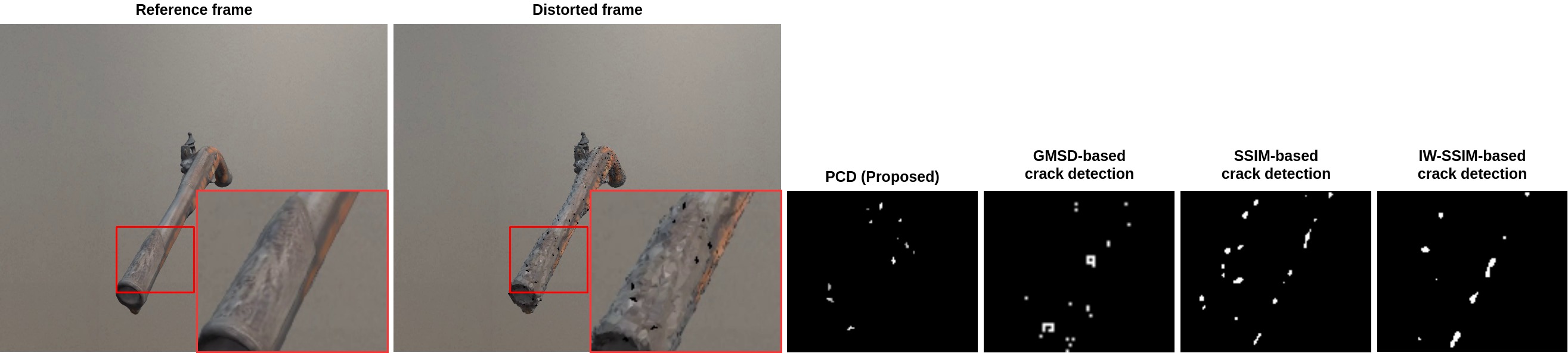}}
    \vspace{5pt}
    \centerline{\includegraphics[width=1.0\textwidth]{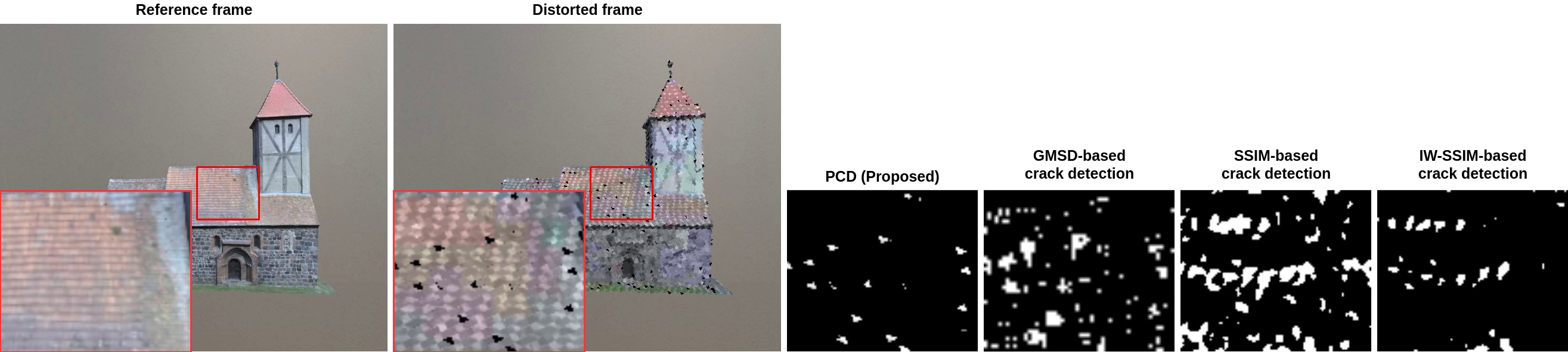}}
    \vspace{5pt}
    \centerline{\includegraphics[width=1.0\textwidth]{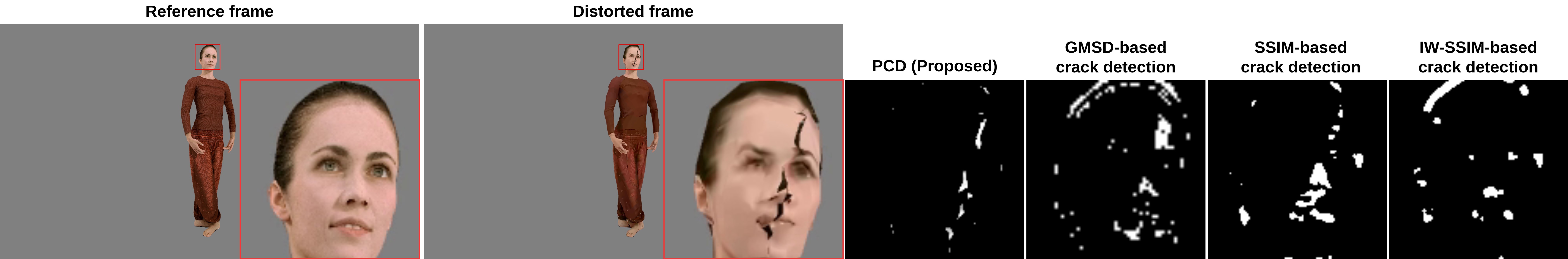}}
    \vspace{5pt}
    \centerline{\includegraphics[width=1.0\textwidth]{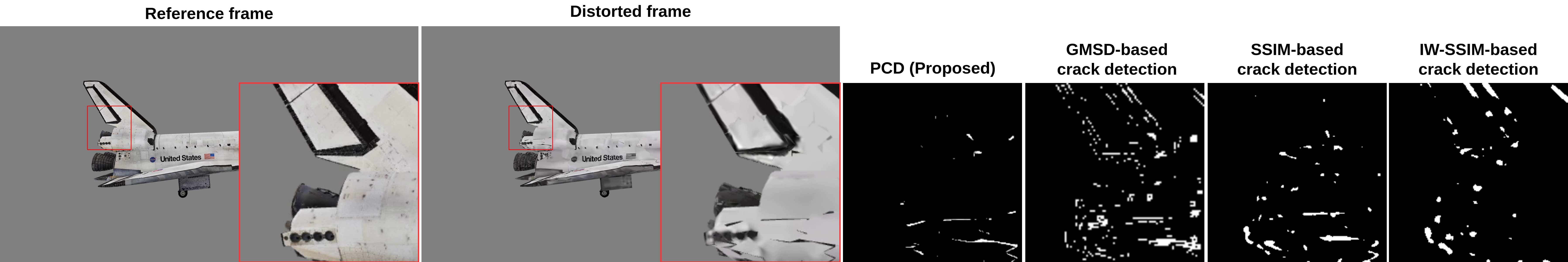}}

    \caption{Crack maps of local windows of four different input pairs of reference and distorted frames obtained from the Nehm{\'e} \emph{et al.} \cite{nehme2023textured} (first two rows) and TSMD \cite{tsmd} (last two rows) datasets. GMSD \cite{gmsd}, SSIM \cite{wang2004image}, and IW-SSIM \cite{iwssim} quality maps are also generated and binarized with optimized thresholds and added as baselines for comparison with the proposed PCD method.}
    \label{fig:crack_map}
\end{figure*}

We propose a novel and effective crack artifact detection and localization method called PCD, drawing upon various HVS theories. Fig.~\ref{fig:crack_detect} summarizes the proposed method. Suppose that $x_i$ and $y_i$ are the $i$-th gray-scale pixels of the corresponding snapshots (frames) of the reference and distorted 3D objects, respectively. They should be taken from the same viewpoint, covering the same angle of the 3D objects. First, the absolute difference between the corresponding pixels of the two frames is computed, i.e. $|d_i| = |x_i - y_i|$. Observe that $|d_i| \in [0, 1], \forall i$. When a pixel $i$ in the distorted frame corresponds to a crack artifact, $|d_i|$ tends to be closer to 1. To avoid magnifying undesired minor artifacts in later stages, we truncate smaller values of $|d_i|$, i.e. if the absolute difference value $|d_i|$ is smaller than a threshold, it is set to 0. The threshold is empirically set to 0.1.
As shown in Fig.~\ref{fig:crack_detect}, the output of this first stage is the truncated absolute difference (TAD) map denoted by $\overline{|d_i|}$.

We then exploit the visual masking effect of HVS \cite{masking_eff, masking_eff2} to modulate the resulting TAD map. We note that crack artifacts occurring in low-contrast (smooth) local regions are more visible and annoying compared to those emerging around the edges and high-contrast (textured) local regions of the 3D object. In other words, textured local regions can provide a masking effect for crack artifacts to some extent \cite{lavoue2009local, wang2012fast, vavsa2012dihedral} following the well-established visual masking effect of HVS. Therefore, we normalize each pixel $i$ of the TAD map by the local contrast of its corresponding pixel in the reference frame, denoted by $\sigma_i$, i.e.
\begin{equation}
    \overline{m}_i = \frac{\overline{|d_i|}}{\sigma_i + C_1},
\end{equation}
where $\boldsymbol{\overline{m}}$ denotes the resulting contrast normalized map, and the constant 
$C_1 = 0.01$ is included to avoid instability in division. To account for local contrast, we adopt the same setting as in the structural similarity (SSIM) approach \cite{wang2004image} by employing a sliding Gaussian window with a standard deviation of 1.5, but a smaller window size of $5\times 5$.

To account for the sensitivity of HVS to high-frequency changes in the appearance of 3D objects, the resulting map is enhanced by a second modulation step with the absolute Laplacian map of the distorted frame denoted by $|l_i|$, i.e.
\begin{equation}
        \widetilde{m}_i = \overline{m}_i\cdot |l_i| = \frac{\overline{|d_i|}\cdot|l_i|}{\sigma_i + C_1},
        \label{eq:init_map}
\end{equation}
where $\boldsymbol{\widetilde{m}}$ denotes the initial crack map, and $\widetilde{m}_i \geq 0, \forall i$. 
This enhancement step is also motivated by the fact that crack artifacts create strong local edges in the distorted frame. 
As such, an unexpected sharp edge in the distorted frame that corresponds to a low-contrast region in the reference frame with a highly different intensity results in a large value, resonating interestingly with our visual definition of the crack artifact.

Finally, as shown in Fig.~\ref{fig:crack_detect}, a truncated sigmoid non-linearity is used to create the final crack likelihood map: 
\begin{equation}
        m_i = f(\widetilde{m}_i) = \left\{\begin{matrix}
               \text{sigmoid}(\frac{\widetilde{m}_i - T_1}{T_1}) & \widetilde{m}_i > T_1\\ 
               0 & \text{otherwise}
            \end{matrix}\right.,
\end{equation}
where by definition $m_i \in \{0\}\cup (0.5, 1], \forall i$. The bigger the value of $m_i$, the more likely the $i$-th pixel is contaminated by the crack artifact. 
The use of sigmoid is motivated by (1) the psychometric saturation effect of the visual system, which states that human observers' ability to discriminate between two different distorted stimuli decreases as the stimuli surpass a threshold of degradation \cite{wang2012fast}; and (2) its capability of generating probability-like outputs for QA tasks \cite{pieapp, yildiz2020machine}.

\subsection{Integration with IQA Models} \label{ssec: int_framework}

\begin{figure*}[t]
    \centering

    \centerline{\includegraphics[width=1.0\textwidth]{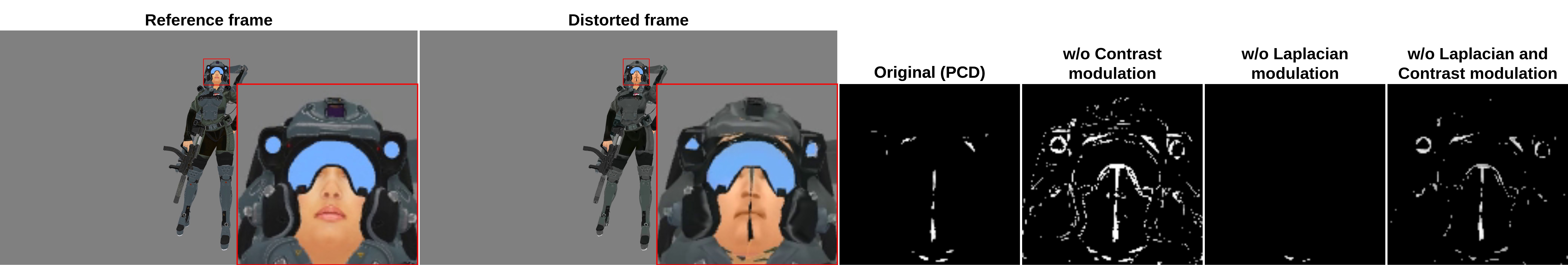}}
    \vspace{5pt}
    \centerline{\includegraphics[width=1.0\textwidth]{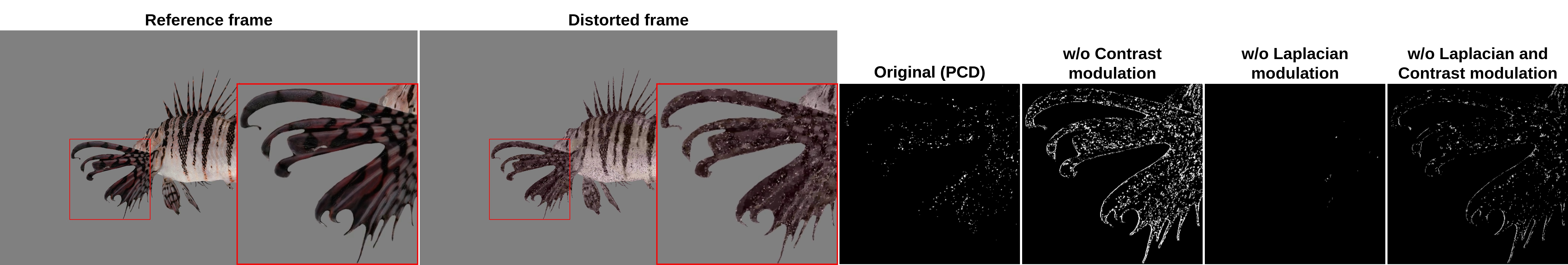}}

    \caption{Ablation study for contrast and Laplacian modulation components of the proposed PCD method for two sets of input pairs of reference and distorted frames obtained from the TSMD dataset \cite{tsmd}.}
    \label{fig:crack_abl}
\end{figure*}

To demonstrate the effectiveness of the proposed PCD method and quantify its performance, we adopt an important feature of HVS and propose a simple yet efficient method to integrate it with existing QA models that produce pixel-level quality maps and observe how it impacts the quality prediction performance of the QA model. 

Given the input reference and distorted frames, we first symmetrically crop both frames around the bounding box of their 3D objects. This step helps eliminate the background pixels irrelevant to the perceived quality. The crack likelihood map $\boldsymbol{m}$ is then incorporated using a weighting scheme, where the weight $w_i$ of the $i$-th pixel is given by
\begin{equation}
    w_i = \frac{1 + C_2}{1 - m_i + C_2},
    \label{eq:wgt_mp}
\end{equation}
where 
a small constant 
$C_2=0.0001$ is included to avoid division instability and to reflect the sensitivity of the weight values to the crack artifacts. By doing so, a non-crack pixel ($m_i = 0$) is mapped to a weight value of 1, whereas a definite crack pixel ($m_i = 1$) results in an extremely larger weight value. The design of the weighting scheme is inspired by the non-uniform perceptual behavior of HVS towards high-quality and low-quality regions of an image. Specifically, low-quality regions have a stronger influence on perceived quality than high-quality regions \cite{pooling_1}. 

In the final step, the weight map is integrated with the quality map of a given QA model, resulting in an overall quality score:
\begin{equation}
    Q = \frac{\sum_{i=1}^{N}w_i q_i}{\sum_{i=1}^{N}w_i}\,,
    \label{eq:enh_q}
\end{equation}
where $q_i$ denotes the $i$-th pixel of the quality map and $N$ is the total number of pixels. The proposed method can be integrated with any given QA model that outputs a perceptual quality map, aiming to enhance their performance by emphasizing the impact of crack artifacts on the perceived quality.

\section{Experiments}

\subsection{Experimental Setup}

To validate the performance of the proposed PCD method and quantify its effectiveness, we run experiments on two large-scale public datasets of 3D textured meshes, which, to the best of our knowledge, are the only datasets that contain crack artifacts: (1) the Nehm{\'e} \emph{et al.} dataset \cite{nehme2023textured}, and (2) the TSMD dataset \cite{tsmd}.

\textbf{The Nehm{\'e} \emph{et al.} dataset} \cite{nehme2023textured} is the largest public dataset of 3D textured meshes. It contains 55 3D source models, each distorted by a mixture of simplification, (vertex position and UV) quantization, texture sub-sampling, and texture compression distortions with different strength levels, leading to a total of 343,750 distorted stimuli \cite{nehme2023textured}. A subset of 3000 stimuli was selected and judged by 4513 subjects through crowdsourcing \cite{nehme2023textured}. The method of subjective study was the double stimulus impairment scale with five scales, and subjects were asked to watch videos of rotating 3D objects before rating them \cite{nehme2023textured}. In our experiments, we use the published videos of 3000 distorted stimuli, their corresponding reference videos, and their reported mean opinion scores (MOSs). The videos are 8 seconds long and in $650\times 550$ resolution with a frame rate of 30 fps \cite{nehme2023textured}.

\textbf{The TSMD dataset} \cite{tsmd} contains 42 3D source models. Similar distortion types as in Nehm{\'e} \emph{et al.} dataset \cite{nehme2023textured} are mixed and applied to each source 3D mesh to generate five distorted meshes per reference mesh, resulting in a total of 210 distorted stimuli. All distorted stimuli were rendered as videos and then judged by 74 viewers through crowdsourcing. The method of subjective study was the double stimulus impairment scale with five scales. Participants were tasked with viewing videos of rotating 3D objects and then providing ratings for them. \cite{tsmd}. In our experiments, we use the 210 distorted videos stimuli, their corresponding reference videos, and their reported MOSs. The videos are 18 seconds long and in $1920\times 1080$ resolution with a frame rate of 30 fps \cite{tsmd}.

To compare the performance of various QA metrics with their enhanced versions, we employ the Spearman rank-order correlation coefficient (SRCC) and the Pearson linear correlation coefficient (PLCC). The PLCC score is obtained after applying a logistic non-linear fitting approach to map predicted quality scores into the MOS space, as recommended by \cite{video2003final} and used in \cite{sheikh2006statistical}.

\subsection{Qualitative Results} \label{ssec: qual_res}

\begin{table*}[t]
    \caption{SRCC ($r_s$) and PLCC ($r_p$) scores of base and enhanced versions of various QA metrics on the Nehm{\'e} \emph{et al.} \cite{nehme2023textured} and TSMD \cite{tsmd} datasets. The bold values indicate the best result in each column for each dataset. The enhanced methods are the results of our proposed integration framework for each base QA model.}
    \centering
    \begin{tabular}{c|c||cc|cc|cc|cc|cc|cc|cc}
        \toprule
        \multirow{2}{*}{\textbf{Dataset}} & \multirow{2}{*}{\textbf{Version}} & \multicolumn{2}{|c}{\textbf{lumaPSNR}} & \multicolumn{2}{|c}{\textbf{SSIM}\cite{wang2004image}} & \multicolumn{2}{|c}{\textbf{MS-SSIM}\cite{wang2003multiscale}} & \multicolumn{2}{|c}{\textbf{IW-SSIM}\cite{iwssim}} & \multicolumn{2}{|c}{\textbf{FSIM}\cite{zhang2011fsim}} & \multicolumn{2}{|c}{$\textbf{FSIM}_C$\cite{zhang2011fsim}} & \multicolumn{2}{|c}{\textbf{Average}} \\
        & & $r_s$ & $r_p$ & $r_s$ & $r_p$ & $r_s$ & $r_p$ & $r_s$ & $r_p$ & $r_s$ & $r_p$ & $r_s$ & $r_p$ & $r_s$ & $r_p$ \\
        \hline
        \multirow{2}{*}{Nehm{\'e} \emph{et al.} \cite{nehme2023textured}} & \textit{Base Method} & 0.469 & 0.485 & 0.333 & 0.372 & 0.465 & 0.480 & 0.564 & 0.581 & 0.559 & 0.579 & 0.556 & 0.577 & 0.491 & 0.512 \\
        & \textit{Enhanced Method} & \textbf{0.657} & \textbf{0.655} & \textbf{0.668} & \textbf{0.676} & \textbf{0.680} & \textbf{0.688} & \textbf{0.703} & \textbf{0.711} & \textbf{0.723} & \textbf{0.729} & \textbf{0.722} & \textbf{0.728} & \textbf{0.692} & \textbf{0.698} \\
        \midrule
        \multirow{2}{*}{TSMD \cite{tsmd}} & \textit{Base Method} & 0.545 & 0.528 & 0.504 & 0.511 & 0.654 & 0.255 & 0.756 & 0.768 & 0.658 & 0.667 & 0.657 & 0.665 & 0.629 & 0.564 \\
        & \textit{Enhanced Method} & \textbf{0.663} & \textbf{0.653} & \textbf{0.699} & \textbf{0.705} & \textbf{0.744} & \textbf{0.760} & 0.757 & 0.767 & \textbf{0.746} & \textbf{0.749} & \textbf{0.746} & \textbf{0.749} & \textbf{0.726} & \textbf{0.731} \\
        \bottomrule
    \end{tabular}
    \label{tab:main_res}
\end{table*}

Fig.~\ref{fig:crack_detect}, \ref{fig:crack_map}, and \ref{fig:crack_abl}  show sample crack maps for various input frames of reference and distorted 3D textured meshes from both datasets. Since, to the best of our knowledge, no other image-based or model-based method has been developed for crack detection of 3D textured meshes, we build our own baseline methods for comparison purposes and include them in Fig.~\ref{fig:crack_map}. Specifically, we adopt quality maps of GMSD \cite{gmsd}, SSIM \cite{wang2004image}, and IW-SSIM \cite{iwssim} QA methods - which can essentially be regarded as artifact localization methods - and select hard thresholds to binarize them into crack and non-crack pixels. Thresholds are carefully chosen by empirical optimization over data samples to generate the best possible results. The proposed PCD method achieves correct detection and localization of crack artifacts in all samples and outperforms all baseline methods that fail to accurately localize and differentiate crack artifacts. Furthermore, samples in Fig.~\ref{fig:crack_map} exhibit other undesired artifacts (e.g. abrupt edges/patterns) that are detected by baseline methods. However, the resulting crack maps only highlight crack artifacts, which shows that the proposed PCD method is capable of differentiating crack from other types of artifacts. Fig.~\ref{fig:crack_map} also illustrates the proposed method's robustness to background and source content variations, showcasing its generalization on two datasets with different backgrounds and source collections.

\subsection{Ablation Study}

We also conduct an ablation study to confirm the functionality of contrast and Laplacian modulation components of the proposed PCD method. Specifically, we explore four versions of the proposed PCD method: (1) the original PCD method; (2) the PCD method without contrast modulation; (3) the PCD method without Laplacian modulation; and (4) the PCD method without contrast and Laplacian modulation. Fig.~\ref{fig:crack_abl} showcases the results for two sets of input reference and distorted frames from the TSMD dataset \cite{tsmd}. As observed, the version without contrast and Laplacian modulation detects all types of distortions and fails to distinguish crack artifacts from other undesired artifacts (e.g. pattern displacement). However, contrast modulation improves outcomes by modulating artifacts according to the visual masking effect of HVS, effectively filtering out undesirable and imperceptible artifacts. Additionally, the application of Laplacian modulation further enhances the results by magnifying crack artifacts according to the sensitivity of HVS to high-frequency details. As seen in Fig.~\ref{fig:crack_abl}, both samples attest to the effectiveness of the proposed PCD method and confirm necessity of contrast and Laplacian modulation components.

\subsection{Quantitative Results}

To quantitatively demonstrate the effectiveness of the proposed PCD method, we quantify the performance gain by the proposed integration method over six well-known base QA models, which include: (1) lumaPSNR, which computes the peak signal-to-noise ratio (PSNR) on the Y channels of reference and distorted frames in the YUV color space; (2) SSIM \cite{wang2004image}; (3) MS-SSIM \cite{wang2003multiscale}; (4) IW-SSIM \cite{iwssim}; (5) FSIM \cite{zhang2011fsim}; (6) and $\text{FSIM}_C$ \cite{zhang2011fsim}. PIQ implementations of IW-SSIM, FSIM, and $\text{FSIM}_C$ were used in our experiments \cite{kastryulin2022pytorch}. For a given QA metric, a single quality score is computed for each pair of (reference and distorted) videos of the Nehm{\'e} \emph{et al.} \cite{nehme2023textured} and TSMD \cite{tsmd} datasets by applying the model to individual frames of the videos and averaging per-frame scores. For better efficiency, highly overlapped frames are ignored, and only one out of every ten consecutive frames is included in the computations, which is sufficient to cover all vertical angles. Two versions of each QA metric are tested on all videos of each dataset: (1) the base method; and (2) the enhanced method which is the result of our proposed integration framework (Section \ref{ssec: int_framework}) for each base QA model. The resulting SRCC ($r_s$) and PLCC ($r_p$) scores are summarized in TABLE \ref{tab:main_res}. We make the following observations. First, the base methods generally exhibit poor performance in all evaluation criteria and both datasets. The QA metrics are developed for natural scenes and their common distortions, while the snapshots of 3D objects are not statistically similar to natural images and are contaminated by a different set of distortions. Second, the proposed integration framework enhances the performance of all base methods on both datasets, regardless of the base model, demonstrating the effectiveness and generalizability of the proposed PCD and integration frameworks. Also, on the Nehm{\'e} \emph{et al.} dataset, with respect to the base models, the integration framework provides $40.9\%$ and $36.3\%$ increases, on average, in terms of SRCC and PLCC scores, respectively. For the TSMD dataset, these average
increases are observed to be $15.4\%$ for SRCC and $29.6\%$ for PLCC.

Furthermore, we conduct an additional study by testing the proposed PCD method as a standalone 3D QA model (without employing any base QA model). Specifically, we assign a crack artifact score (CAS) to a given pair of (reference and distorted) frames by averaging the pixel values of the computed crack map. The larger the CAS, the lower the quality of the frame.
When tested on all videos of the Nehm{\'e} \emph{et al.} dataset, CAS achieves SRCC and PLCC scores of $0.665$ and $0.655$, respectively; while on the TSMD dataset, it attains SRCC and PLCC scores of $0.675$ and $0.651$, respectively. Interestingly, CAS as a standalone QA model outperforms all QA base models on the Nehm{\'e} \emph{et al.} dataset and most of them on the TSMD dataset, validating the effectiveness of the proposed PCD method and highlighting the major influence of crack artifacts on perceived image quality.

\subsection{Runtime Analysis}

\begin{table}
    \caption{Runtime analysis of the proposed PCD and baseline methods for sample input frames of $650\times 550$ and $1920\times 1080$ resolution. CD stands for crack detection.}
    \centering
    \begin{tabular}{c|c|c}
        \toprule
        \multirow{2}{*}{\textbf{Operation}} & $\boldsymbol{650\times550}$ & $\boldsymbol{1920\times1080}$ \\
        & \textbf{Resolution} & \textbf{Resolution} \\
        \hline
        PCD (proposed) & 0.009s & 0.060s \\
        GMSD-based \cite{gmsd} CD & 0.004s & 0.018s \\
        SSIM-based \cite{wang2004image} CD & 0.029s & 0.238s \\
        IW-SSIM-based \cite{iwssim} CD & 0.055s & 0.423s \\   
        \bottomrule
    \end{tabular}
    \label{tab:runtime}
\end{table}

Finally, we perform a runtime analysis for the proposed PCD method. Table \ref{tab:runtime} presents the PCD's crack map computation time for input video frames of $650\times 550$ (Nehm{\'e} \emph{et al.} dataset \cite{nehme2023textured}) and $1920\times 1080$ (from TSMD dataset \cite{tsmd}) resolution and compares it with the processing time needed for the baseline crack detection methods as introduced in Section \ref{ssec: qual_res}). The analysis was conducted on a computing platform equipped with an Intel Core i7-12700K CPU. As we can see, the proposed PCD method is highly efficient and provides a real-time computation of the crack maps for all frames of a video of $650\times 550$ resolution with a 100 fps frame rate. Near real-time performance can also be achieved in videos of $1920\times 1080$ resolution with 25 fps frame rate. Furthermore, we can observe that the crack map computation runs much faster than the baseline SSIM-based and IW-SSIM-based crack detection methods and slightly slower than the baseline GMSD-based crack detection method. Given its efficiency, PCD can be applied across various stages of a 3D mesh supply pipeline to promptly identify crack artifacts in distorted 3D textured meshes.

\section{Conclusion}
We propose PCD, a novel Perceptual Crack Detection method for 3D textured meshes. The proposed method operates on a pair of input snapshots of distorted and reference 3D objects and takes advantage of HVS characteristics and visual characteristics of crack artifacts to generate a crack likelihood map that highlights contaminated pixels. Additionally, to quantitatively validate the effectiveness of the proposed PCD method, we propose a simple yet efficient framework for integration of the crack map with existing QA models to boost their performance in 3D QA tasks. Experiments on large-scale public datasets of 3D textured meshes demonstrate the efficiency and effectiveness of the proposed PCD and integration frameworks.







\bibliographystyle{IEEEtran}
\bibliography{refs.bib}
%
%
%

\end{document}